\documentclass[preprint]{elsarticle}
\usepackage[utf8]{inputenc}
\usepackage{natbib}
\usepackage{float}
\usepackage{amssymb}
\usepackage{amsmath}
\usepackage{booktabs}
\usepackage{tabularx}
\usepackage{multirow}
\usepackage{todonotes}
\usepackage{caption}
\usepackage{subcaption}
\usepackage{lineno}

\cortext[cor1]{Corresponding author}
\address[1]{Technical University of Munich (TUM), Data Science in Earth Observation, Arcisstr.~21, 81333~Munich, Germany}
\address[2]{German Aerospace Center (DLR), Remote Sensing Technology Institute, Münchener~Str.~20, 81345~Wessling, Germany}

\begin{document}

\begin{frontmatter}

\title{Using Social Media Images for Building Function Classification}

\author[1]{Eike Jens Hoffmann}
\author[2]{Karam Abdulahhad}
\author[1,2]{Xiao Xiang Zhu \corref{cor1}}
\ead{xiaoxiang.zhu@dlr.de}
\date{February 2022}

\begin{abstract}
Urban land use on a building instance level is crucial geo-information for many applications, yet difficult to obtain. 
An intuitive approach to close this gap is predicting building functions from ground level imagery. 
Social media image platforms contain billions of images, with a large variety of motifs including but not limited to street perspectives. To cope with this issue this study proposes a filtering pipeline to yield high quality, ground level imagery from large social media image datasets. The pipeline ensures that all resulting images have full and valid geotags with a compass direction to relate image content and spatial objects from maps.

We analyze our method on a culturally diverse social media dataset from Flickr with more than 28 million images from 42 cities around the world. The obtained dataset is then evaluated in a context of 3-classes building function classification task. The three building classes that are considered in this study are: commercial, residential, and other. Fine-tuned state-of-the-art architectures yield F1-scores of up to 0.51 on the filtered images. Our analysis shows that the performance is highly limited by the quality of the labels obtained from OpenStreetMap, as the metrics increase by 0.2 if only human validated labels are considered. Therefore, we consider these labels to be weak and publish the resulting images from our pipeline together with the buildings they are showing as a weakly labeled dataset.
\end{abstract}


\end{frontmatter}


\section{Introduction}
\label{sec:introduction}
Building function classification is the task of automatically identifying the settlement type of a given building, e.g., is it a residential or commercial building? On the one hand, the task operates on a fine-grained level, i.e., single building level. On the other hand, it is an essential task for urban planning and decision making, especially in the big and dynamic cities. Historically, this task was manually managed by cadastral offices. However, this process is high resource consuming, and it could not catch up with the size and the speed of development of the modern cities. To cope with this issue, automatic methods are applied, where they mainly consume air-view images, such as aerial or satellite images \cite{huang2018urban, zhang2019joint}. Although this kind of data is of high quality, it has inherent ambiguities from a nadir view looking at rooftops.

\subsection{Motivation}
Social media data, on the other side, is ubiquitous, cheap and easy to collect. It has become an essential and valuable source of information for many applications and scenarios \cite{kruspe2021changes}. In our task, social media data shows promising features to replace classic air-view sources of data. First, it offers a ground-level view, which means a finer-grained and a different perspective source of data. Second, it is a more up-to-date source of information, or it is even a real-time source of information. Third, it is a gigantic source of cheap data. The only restriction in our scenario is that we need geo-tagged social media data. Fortunately, this is the case for a considerable share of data that is coming from social media channels such as Twitter or Flickr. For example, around 1\% of all tweets are geo-tagged \cite{lsx2013}, i.e., given that around 500M\footnote{https://www.internetlivestats.com/twitter-statistics/} tweets are published per-day, 5M among them are geo-tagged. Flickr does not disclose its photo statistics in detail, but announces having \textit{billions of photos already online}\footnote{https://www.flickr.com/jobs}. By aligning geo-tagged social media content with open Volunteered Geographical Information (VGI) systems, such as OpenStreetMap (OSM), we could decode social media posts (e.g., tweets, images, etc.) to certain places on earth, and, hopefully, to certain buildings. However, one should not take social media as a no-cost source of information. One should be careful when dealing with social media as a main source of information, where it is a noisy and uncontrolled source of data. In addition, it is a sparse source, where not all spots on earth are equally covered by social media. For example, Flickr photos are mainly coming from city centers and hotspots.

\subsection{Related Work}

Urban land use classification is a challenging task: no matter on which spatial level it is done, there are inherent ambiguities. On the most fine-grained level, building instances, it is often hard to decide which land use a building represents. Especially in dense, urban centers mixed uses are the most dominant class.

Several studies investigated the feasibility of building facade images to address this problem. There are two main sources for such ground level image data: first, commercial street view data like Google Street View or Mapillary and second, social media platforms like Facebook, Instagram, or Flickr.

Especially Google Street View is a preferred source for this task as its data is accessible using an API enabling the user to define position, heading, pitch, and field-of-view. Additionally, Google has its own, standardized hardware to capture street view images and a tailored image processing pipeline to generate high-quality imagery on scale.

In combination with Google Places data, Google Street View data allows fine-grained store classification \cite{movshovitz2015ontological}. This work builds upon Google Map Maker ontology and a GoogLeNet architecture trained on global sample of Google Street view.

Access to the Google Places is limited for research outside of Google and moreover, Google Places focuses on points of interest (POIs) and does not include data about residential buildings. As an alternative, building footprints from OpenStreetMap (OSM) can have semantic data as well, including details about building functions. This information can be used to label buildings shown in Google Street View images and hence, provides an additional way to predict land use on a building instance level \cite{kangBuildingInstanceClassification2018}.

The comprehensive coverage of buildings by Google Street View allows to have multiple images from different perspectives for a single building. This data richness can be used in a multi-modal architecture to include information from different sides while obtaining the labels from OSM \cite{srivastava2020fine}.

Beyond land use classification information encoded in Google Street View images can be used to infer socioeconomic characteristics \cite{gebru2017using} or to map urban green in terms of tree detection and positioning \cite{laumer2020geocoding}.

However, the terms of service of Google Street View prohibit scraping, downloading, or storing images obtained using the API. This legal constraint limits the applicability of Google Street View data in research projects and requires to analyze other sources of data, e.g. social media image platforms. While Facebook and Instagram do not open their data for such purposes, Flickr turned out to be a valuable image source as they provide an easy accessible API and encourage their users to share photos with creative commons license.

While early works on land use classification with Flickr images used bag-of-visual-word features for classification \cite{leung2012exploring}, more recent studies benefited from advancements in computer vision with CNNs and proposed land use classification using a scene and a object detection stream in parallel \cite{zhu2019fine}.

On a larger spatial scale Flickr has been used for mapping and understanding landscape aesthetics, either manually \cite{langemeyer2018mapping} or based on CNNs \cite{salem2020learning, havinga2021social}. Another field of application is flood level estimation. By formulating this problem as an object detection task with Mask R-CNN it has been shown that these images help to predict discrete levels of flooding \cite{chaudhary2019flood}.

If social media images are used for a specific application, it is crucial to deal with huge variation in motifs and scenes. Other data sources with a dedicated purpose but limited spatial extent can be a better option in some cases. For example, images from Geograph project\footnote{https://www.geograph.org.uk/} are captured in a systematic way to cover Great Britain and Ireland. Its aim is to have at least one representative image for every square kilometer on both islands. These images can be used for predicting urban land use in London with object bank features \cite{li2010object, fang2018urban}.

Apart from Flickr, there is also Twitter as social media data source providing geo-located information with textual features. Although Twitter restricted their geo-tagging feature in June 2019 it is still a valuable source of geospatial data \cite{kruspe2021changes}. To predict building functions it can be sufficient to have a set of geo-tagged tweets and build a classifier using their metadata \cite{huang2018classification}. As tweets contain mainly text, the inherent linguistic features have also shown potential to help in urban land use classification on a building instance level \cite{haberle2019building} as well as on a venue level \cite{terroso2020land}.

Furthermore, geo-located Twitter data reveal patterns in language use and provides insights into socioeconomic factors when related with demographics \cite{bokanyi2016race}. When used in combination with Flickr data correlation between socioeconomic factors and parks visits show up \cite{hamstead2018geolocated}.

\subsection{Contribution}
In this paper, we tackle the problem of building function classification using Flickr images. To make Flickr images useful for our task, they should first pass through a rigid filtering pipeline to eliminate noisy, irrelevant, and non geo-tagged photos. After that, a Convolution Neural Network (CNN) is finetuned for a multi-class classification downstream task. In this study, we mainly consider three classes of buildings from OSM, namely residential, commercial, and other. The main contribution of this paper can be summerized in the following points:
\begin{itemize}
    \item Building function classification using weakly labeled Flickr images.
    \item A content-based automatic filtering pipeline to eliminate irrelevant and noisy Flickr photos from large-scale and real-world datasets
    \item A human-validated subset of Flickr photos for testing.
\end{itemize}

\section{Methodology}
\label{sec:methodology}
Our method uses social media images to classify buildings settlement types. We follow a content-based approach, which is able to identify the main visual patterns for each class. 

\subsection{Homogenizing OpenStreetMap Building Labels}
OpenStreetMap (OSM) is a Volunteered Geographic Information (VGI) platform meaning that users contribute mapping data in a Wikipedia-like style. OSM provides guidelines how this data should be structured and semantically enriched, but there is no strict enforcement. Therefore, tags for buildings are optional, just the building footprint coordinates are mandatory if added to OSM.
OSM's guidelines specify three different tags that can be added for indicating a building function: \textit{building}, \textit{amenity}, and \textit{shop}.

To summarize the information from all three tags, we use a mapping scheme that assigns each possible value according to OSM's guidelines of each tag to one of \textit{commercial}, \textit{residential}, and \textit{other}. If more than one of these tags, \textit{building}, \textit{amenity}, and \textit{shop}, is present, we make sure that they do not disagree. In case of disagreement, the building is not mapped to any of the classes. If there is only one tag or all available ones agree on the same mapped class, then this building gets this class.

\subsection{Social Media Image Filtering Pipeline}
Social media images cover different content and motifs, including but not limited to photography, digital art, cartoons. However, given a task like building function classification, most of these images are not helpful to solve the task. For our task, an image must have three features:

\begin{enumerate}
  \item Shows a building
  \item Has a valid geotag
  \item Has a known compass orientation
\end{enumerate}

A filtering pipeline needs to identify all images fulfilling these three criteria in a social media image dataset. Additionally, it must account for big data to work on datasets with millions of images.

\begin{figure}[H]
  \centering
  \includegraphics[width=\textwidth]{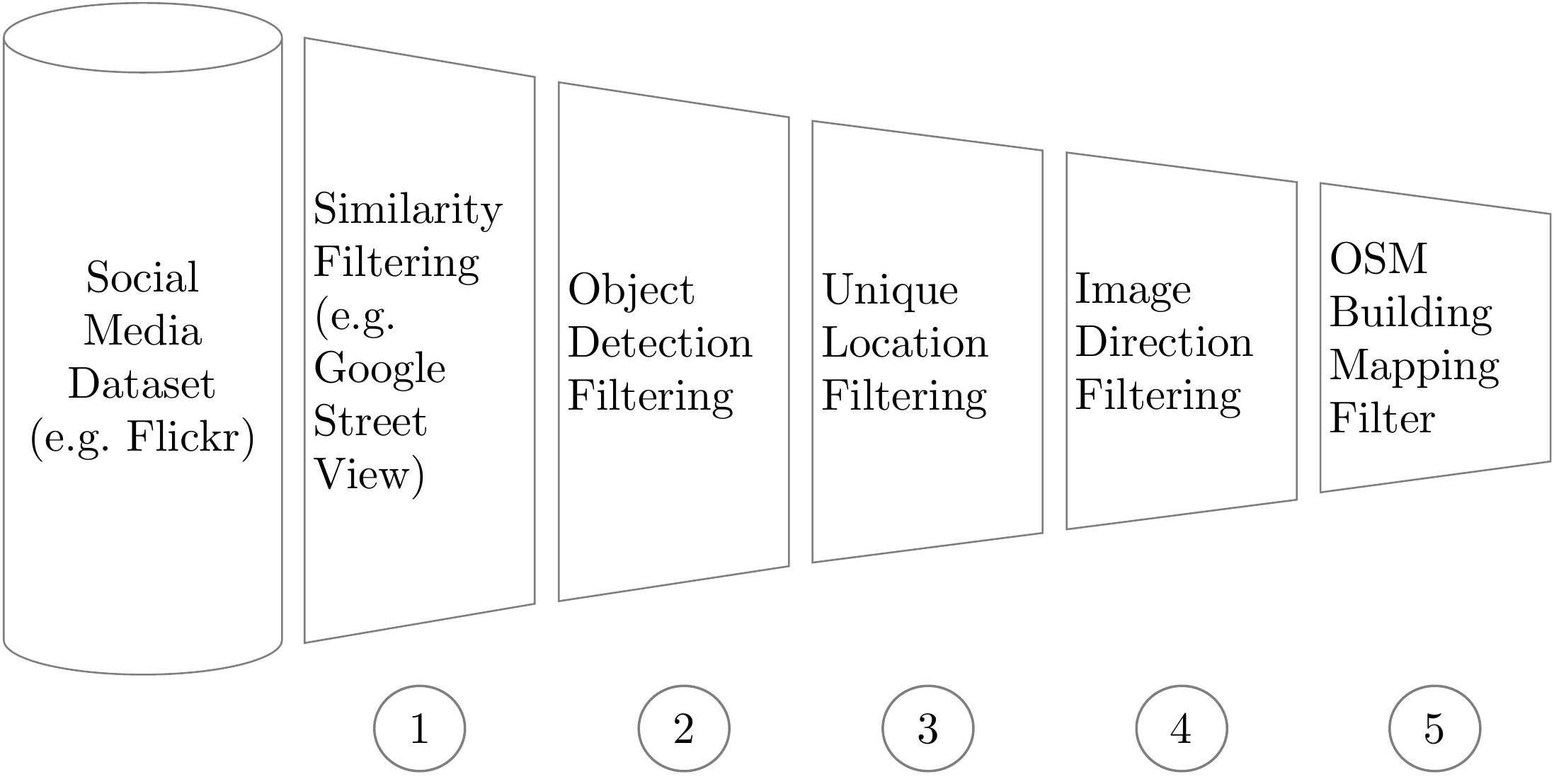}
  \caption{Filter pipeline for extracting Street View-like images from Flickr image database}
  \label{fig:flickr_filter_pipeline}
\end{figure}

Figure~\ref{fig:flickr_filter_pipeline} shows the pipeline used in this study. It consists of five steps, starting with Google Street View similarity filtering and object detection filtering. These two steps together ensure that the first criterion is matched. We validate geotags in the next two steps: first, with an heuristic that discards images whose location is not unique. If there is another image at exactly the same position it is very likely that the geotag was manually edited. Second, we download the metadata for each remaining image and check if it contains a compass orientation. This serves as a stricter check for the second criterion and ensures the last criterion as well. Finally, we use the geotag including the compass direction for spatial referencing with OSM buildings.

\subsubsection{Google Street View Similarity Filtering}
This first step is a coarse filtering step aiming at finding images that are potentially helpful for building function classification. Previous studies showed the relevance of façade images to predict building functions \cite{srivastava2020fine, kangBuildingInstanceClassification2018}. Therefore, this step is formulated as an image retrieval problem with a sample of Google Street View images as seed dataset $S$ and a social media dataset $D$.

Features from deep neural networks are well suited to find structurally similar images. As they aggregate information with every layer, the final layers of a network are an abstract representation of the whole image. For example, the deep features of VGG16 \cite{simonyanVeryDeepConvolutional2015} have been successfully applied in different domains for image retrieval \cite{liuIntelligentSecureContentBased2019,wangBeautyProductImage2018,haImageRetrievalUsing2018,geExploitingRepresentationsPretrained2018}.

In this study, features are taken from the last hidden layer of a VGG16 network trained on ImageNet \cite{russakovskyImageNetLargeScale2015}. This process yields feature vectors $v \in \mathbb{R}^{4096}$. To assess similarity between pairs of images $i_1, i_2$, the cosine similarity $s_{cos}$ is calculated based on the feature vectors $v_1, v_2$:

\begin{equation}
  s_{cos}(v_1, v_2)=\frac{v_1 v_2^T}{\left\| v_1 \right\| \left\| v_2 \right\|}
\end{equation}

For efficient calculation, the features for all images of the seed dataset are calculated beforehand. Then, the features for all social media images are computed batch-wise and we calculate the pair-wise cosine similarity between the batch and the seed dataset. For each social media image in the batch we save the maximum similarity against all seed images, called the similarity parameter $p_{sim}$:

\begin{equation}
  p_{sim}(v_s)=\max\left(\left\{s_{cos}(v_1, v_s), ..., s_{cos}(v_n, v_s)\right\}\right)
\end{equation}

A threshold $t_{sim}$ is set as a minimum similarity value and all social media images with $p_{sim} < t_{sim}$ are discarded.

\subsubsection{Object Detection Filtering}
The previous step is a fast check for structural similarity to a given seed dataset, but does not ensure that the social media images actually contain a building façade. Therefore, this step uses an object detection algorithm to find all objects in the images that passed the previous filter.

Applying the object detection algorithm yields a list of objects for each image. If this list contains either a \textit{house} or a \textit{building} it is a candidate for passing this filter. Each detected object comes with a size relative to the image and a confidence score. Based on these variables, there are two thresholds for adjusting if a candidate image passes the filter: $t_{size}$ and $t_{score}$. Only if there is a building or a house with a size parameter $p_{size}$ that is larger than $t_{size}$ and has a confidence parameter $p_{score}$ higher than $t_{score}$ the image is passed to the next step.

\subsubsection{Unique Location Filtering}
The previous steps confirmed that the image content is relevant for the given task. Now, this step focuses on the geotag of the image. Geotags can be created in two different ways: either automatically by a GPS sensor of the camera or manually by the user.

This filter is a heuristic to identify images that have been manually tagged. If users have to pick locations of images by hand they tend to do it batch-wise, tagging multiple images at the same place. Otherwise, images  tagged using a GPS sensor will have small differences in the position even if the photographer has not moved. GPS sensors constantly update their location estimate based on how many GPS satellites signals are available. Therefore, having two images with exactly the same position is a strong indicator that their geotag has not been measured by a GPS sensor, but manually added. In such cases, there is no compass orientation in the EXIF data and hence, this image can be omitted for the subsequent step.

More formally, an image $i$ from a set of images $I$ with location $l(i)$ passes this filter if

\begin{equation}
\forall j \in I, j \neq i \nexists l(i) = l(j)
\end{equation}

A note on implementation: to make this step computationally efficient, a sequential scan for each image is not feasible. If naïvely done, the geotag for each image needs to be compared with all geotags in the database. A geospatial index decreases the necessary checks by excluding geotags that are far away. Using a R-tree \cite{guttman1984r} allows to find the images in the very close neighborhood and a subsequent check on true equality is performed only on the geotags of these images.

\subsubsection{Image Direction Filtering}
This step is based on metadata of images, so called EXIF data. EXIF is a standard established by the Camera and Imaging Products Asscociation (CIPA) and the Japan Electronics and Information Technology Industries Asscociation (JEITA) \cite{cameraimagingproductsassociationExchangeableImageFile2019}. It defines fields for saving details about images including date and time of capturing, camera model, and camera settings. Moreover, it specifies how data from GPS sensors can be incorporated. This data can be a position of longitude and latitude as well as a compass direction.

For our pipeline, we assume that the social media database does not contain the original images including the EXIF metadata, but only a downsampled variant without original EXIF data. Therefore, we download the EXIF data for all images passing the previous filters as an intermediate step. Once all EXIF data are available, this step checks if the tag \textit{GPSImgDirection} is present and rejects all images that do not have this tag.

Knowing the position where an image was taken is a necessary pre-condition, but only in combination with the compass direction a geospatial reference becomes feasible. Both information together allows computing a line-of-sight, which is crucial for the next step.

\subsubsection{OSM Reference Building Filtering}
This final step establishes a connection between buildings shown in an image and their representations in OSM. We use the position and the compass orientation to create a line-of-sight. All buildings polygons intersecting  the line-of-sight are possible candidates for the building shown in an image. We select the building with the closest distance to the position as the reference building in the picture and set this as parameter $p_{dist}$. Based on this parameter we add a forth threshold $t_{dist}$ to analyze the effect of the distance.

For evaluation, we add another filtering step that discards all images, which are assigned to a building without a semantic label.

\subsubsection{Filtering Pipeline Summary}
Having the pipeline in this order enables a content-first strategy while keeping the computational effort low. Additionally, the number of hyperparameters is small with four thresholds:

\begin{enumerate}
    \item minimum seed similarity $t_{sim}$
    \item minimum object size $t_{size}$
    \item minimum object score $t_{score}$
    \item maximum building distance $t_{dist}$
\end{enumerate}

\subsection{Fine-tuning CNN Architectures for Building Function Classification}
To classify buildings shown in the social media images we fine-tune six state-of-the-art CNN architectures (DenseNet \cite{huangDenselyConnectedConvolutional2018},
InceptionV3, 
\cite{szegedyRethinkingInceptionArchitecture2015}
MobileNetV2, 
\cite{sandlerMobileNetV2InvertedResiduals2019}
ResNetV2, 
\cite{heIdentityMappingsDeep2016}
VGG16,
\cite{simonyanVeryDeepConvolutional2015}
Xception, 
\cite{cholletXceptionDeepLearning2017}). Starting with weights from ImageNet \cite{russakovskyImageNetLargeScale2015} we applied a two step approach to adapt the models for building function classification \cite{hoffmannModelFusionBuilding2019}. We start with ImageNet models without the classification head and add a dense layer with three outputs to predict each of the aforementioned homogenized OSM mapping scheme: \textit{commercial}, \textit{other}, and \textit{residential}. Please note that we fine-tune the models on the Google Street View seed dataset and use social media images only for inference to predict building functions.

As a first step, all layers are frozen and only the new, randomly initialized layer is trained with a learning rate of $lr=10^{-4}$ for at most 16 epochs. Hence, the new layer is adapted to the existing weights and there is no risk of collapsing weights when trained on the full network. To prevent overfitting, a checkpoint mechanism makes sure that after training the model with the lowest validation loss is restored and used for the next step.

After convergence of the newly added layer, the whole model is set as trainable and fine-tuned in a second step with a learning rate of $lr=10^{-5}$. Again, we apply the checkpoint mechanism and create the final model based on the one with the lowest validation loss during training of 16 epochs.

\subsection{Human Label Validation}
To validate the labels obtained from OpenStreetMap buildings, we asked a group of humans to verify labels given to an image. Given a question if an image contains a \textit{commercial}/\textit{other}/\textit{residential} building, they had to choose between three options: \textit{yes}, \textit{unsure}, or \textit{no}. If \textit{no} was selected, users were asked for the correct label in their opinion.

As our classification scheme does not include mixed use labels, we asked our voters to opt for \textit{unsure} if no clear label could be assigned. To make the votes more reliable, our system requested three votes from different humans for each image. Once an image received three votes, it was not shown to any other user again. The users were not restricted in the number of images to vote on. 

\section{Experiments}
We first introduce the two datasets used in this study and continue with describing the results of the different filtering steps. Moreover, we show the results of a Google Street View trained model on filtered Flickr images and dive deeper into the prediction performance by including results from the human validation setup. 

\subsection{Datasets}
We evaluate our method using two datasets: First, a sample of Google Street View (GSV) images featuring buildings with known functions, and second, a Flickr image dataset captured in 42 cities distributed globally.

\subsubsection{Google Street View Dataset}
The Google Street View dataset consists of 43,392 building facade images, distributed to 14,512 commercial, 14,184 other, and 14,696 residential buildings. We apply a faster R-CNN \cite{renFasterRCNNRealTime2016} trained on OID v4 \cite{kuznetsovaOpenImagesDataset2020} to detect objects on all images and discard all images that do not show a \textit{building} or \textit{house}. This combination of architecture and dataset has the best trade-off between accuracy and speed \cite{huangSpeedAccuracyTradeoffs2017}. This yields a refined dataset of 7,698 images (2,743 commercial, 2,333 other, and 2,622 residential).

This Google Street View dataset is used in two ways: first, as a seed dataset for finding structurally similar images in the social media dataset, and second, for fine-tuning state-of-the-art CNN architectures on the given task.

\subsubsection{Flickr Social Media Dataset}
We collected Flickr image data in 42 cities across the globe to cover different cultures and continents. The images were obtained by querying the Flickr API with small random bounding boxes inside these regions of interest. With this approach we harvested 28,818,438 images.

\begin{table}[htbp]
\centering
\begin{tabular}{lr}
\toprule
        City &  \#Images \\
\midrule
   Amsterdam & 1,147,657 \\
     Beijing &   358,393 \\
      Berlin &   929,508 \\
       Cairo &   110,297 \\
   Cape Town &   165,848 \\
    Changsha &     8,051 \\
     Cologne &   610,185 \\
    Dongying &       153 \\
  Guang Zhou &    81,585 \\
   Hong Kong &   964,733 \\
   Islamabad &     9,779 \\
    Istanbul &   259,141 \\
     Jakarta &   204,792 \\
       Kyoto &   668,547 \\
      Lisbon &   463,992 \\
      London & 3,978,803 \\
 Los Angeles & 1,979,163 \\
      Madrid &   709,029 \\
   Melbourne &   661,921 \\
       Milan &   735,996 \\
      Moscow &   569,651 \\
      Mumbai &   140,495 \\
      Munich &   391,798 \\
     Nairobi &    32,262 \\
     Nanjing &    24,411 \\
New York City & 2,351,955 \\
       Paris & 1,344,000 \\
     Qingdao &    11,960 \\
Rio De Janeiro &   425,874 \\
        Rome &   570,033 \\
San Francisco & 1,744,662 \\
    Santiago &   269,656 \\
    Sao Paulo &   729,197 \\
    Shanghai &   330,229 \\
    Shenzhen &    51,893 \\
      Sydney &   730,823 \\
      Tehran &    21,999 \\
       Tokyo & 1,361,486 \\
   Vancouver &   834,973 \\
Washington D.C. & 1,139,602 \\
       Wuhan &    25,754 \\
      Zurich &   288,903 \\
\bottomrule
\end{tabular}
\caption{Number of Flickr images per city}
\label{tab:flickr_images_per_city}
\end{table}

Table~\ref{tab:flickr_images_per_city} shows the number of images per city. The number of images per city correlates with the user distribution of Flickr, so we see the highest number of images in London ($\sim$4.0M images). Second, there is New York City with $\sim$2.3M images and third, Los Angeles with $\sim$1.9M images. Except for Dongying, we found more than 5,000 images in every city. There is evidence that Dongying is a ghost city, meaning that the housing capacity outnumbers the number of inhabitants by far \cite{leichtleHasDongyingDeveloped2019}.

\subsection{Filtering Pipeline Results}
\begin{figure}[htbp]
  \centering
     \begin{subfigure}[b]{0.48\textwidth}
         \centering
         \includegraphics[width=\textwidth]{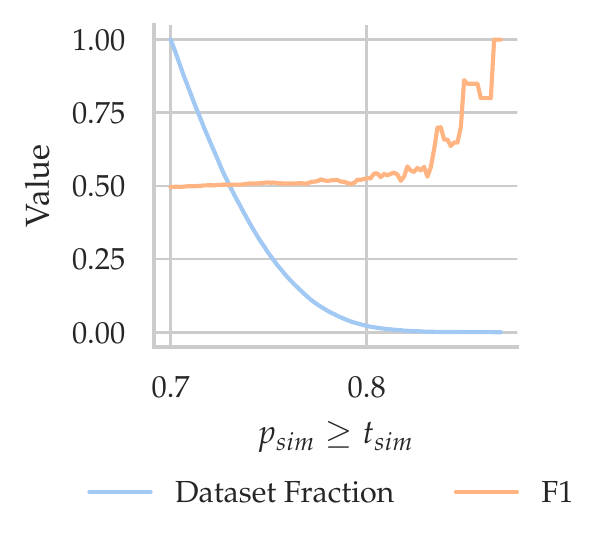}
         \caption{F1-score and dataset size as a function of the similarity threshold $t_{sim}$, while $t_{score}=0.3$, $t_{size}=0.2$, and $t_{dist}=250$}
         \label{fig:analysis_f1_score_t_sim}
     \end{subfigure}
     \hfill
     \begin{subfigure}[b]{0.48\textwidth}
         \centering
         \includegraphics[width=\textwidth]{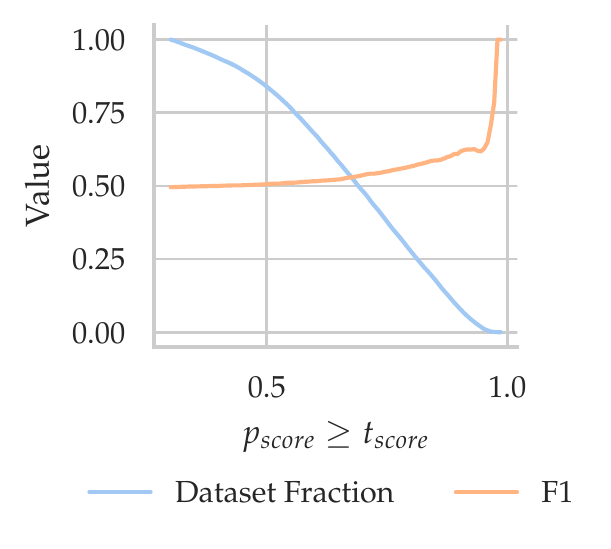}
         \caption{F1-score and dataset size as a function of the object detection score $t_{score}$, while $t_{sim}=0.7$, $t_{size}=0.2$, and $t_{dist}=250$}
         \label{fig:analysis_f1_score_t_score}
     \end{subfigure}
     
     \begin{subfigure}[b]{0.48\textwidth}
         \centering
         \includegraphics[width=\textwidth]{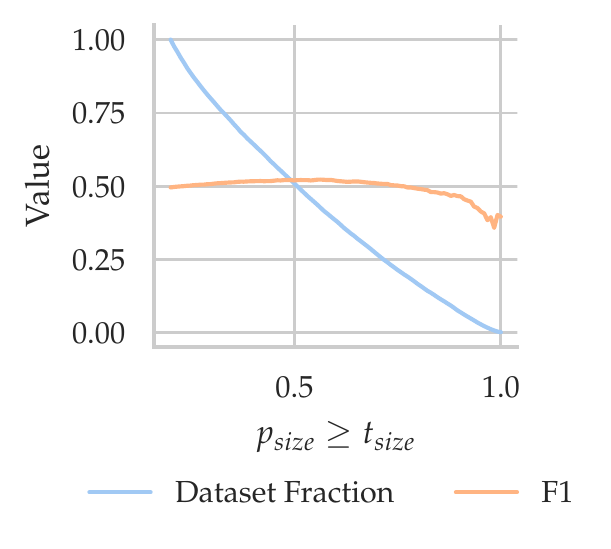}
         \caption{F1-score and dataset size as a function of the relative size of a detected building $t_{size}$, while $t_{sim}=0.7$, $t_{score}=0.3$, and $t_{dist}=250$}
         \label{fig:analysis_f1_score_t_size}
     \end{subfigure}
     \hfill     
     \begin{subfigure}[b]{0.48\textwidth}
         \centering
         \includegraphics[width=\textwidth]{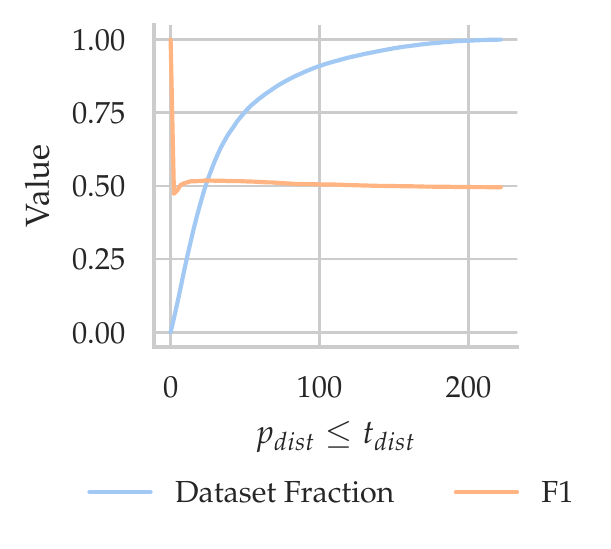}
         \caption{F1-score and dataset size as a function of the distance to the next labeled building $t_{dist}$, while $t_{sim}=0.7$, $t_{score}=0.3$, and $t_{size}=0.2$}
         \label{fig:analysis_f1_score_t_dist}
     \end{subfigure}     
  \caption{Effect of filtering pipeline parameters on prediction results and remaining dataset size}
  \label{fig:filter_pipeline_parameters_f1_score_analysis}
\end{figure}

We evaluate our pipeline end-to-end by analyzing the effects of the four hyperparameters on the F1-score. For an architecture independent evaluation we calculated the mean probability vectors of all six models for each image. Using mean probability vectors of six models eliminates artefacts from single models and allows more general conclusions. Figure~\ref{fig:filter_pipeline_parameters_f1_score_analysis} shows the F1-scores and the remaining dataset size as functions of a threshold. Computing the F1-score requires working on the final output of the pipeline with each image being assigned to an individual building. Hence, the complete dataset of 100\% is based on 26,381 images, 8,070 labeled as commercial, 9,171 labeled as other, and 9,140 labeled as residential.

\begin{figure}[H]
  \centering
  \includegraphics[width=\textwidth]{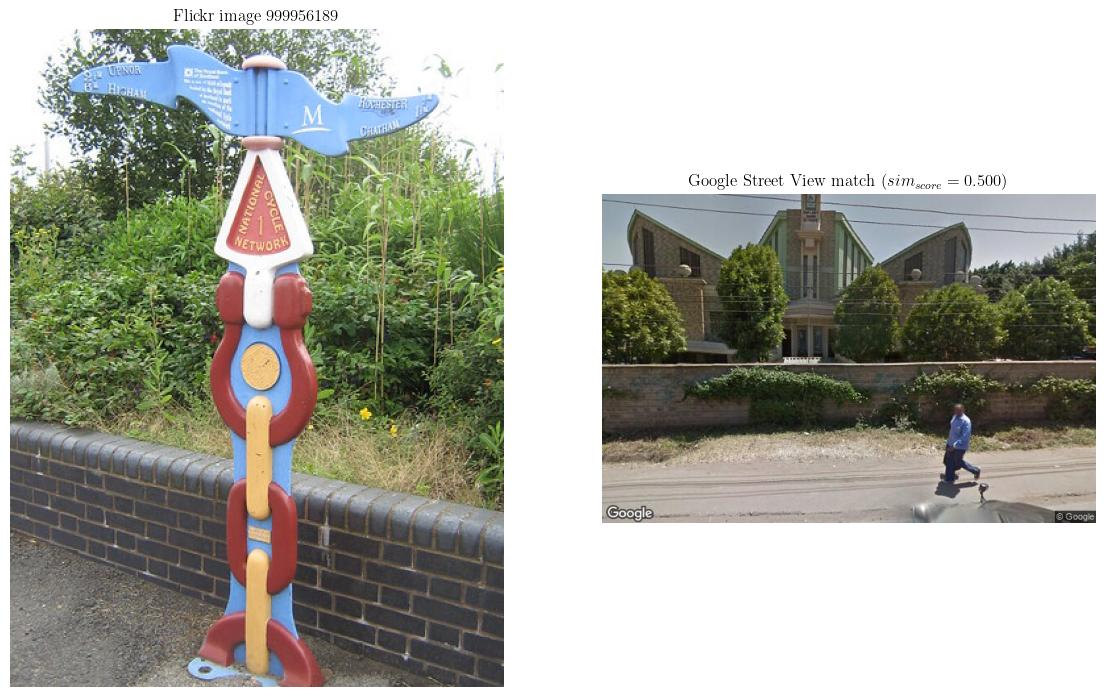}
  \caption{Sample for Flickr image having a $p_{sim}=0.50$}
  \label{fig:sample_flickr_image_0.50_similarity}
\end{figure}

\begin{figure}[H]
  \centering
  \includegraphics[width=\textwidth]{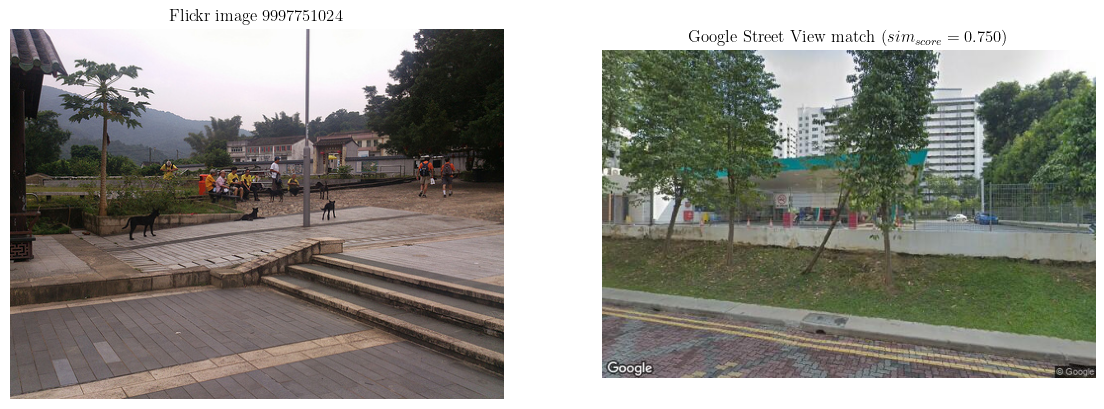}
  \caption{Sample for Flickr image having a $p_{sim}=0.75$}
  \label{fig:sample_flickr_image_0.75_similarity}
\end{figure}

\begin{figure}[H]
  \centering
  \includegraphics[width=\textwidth]{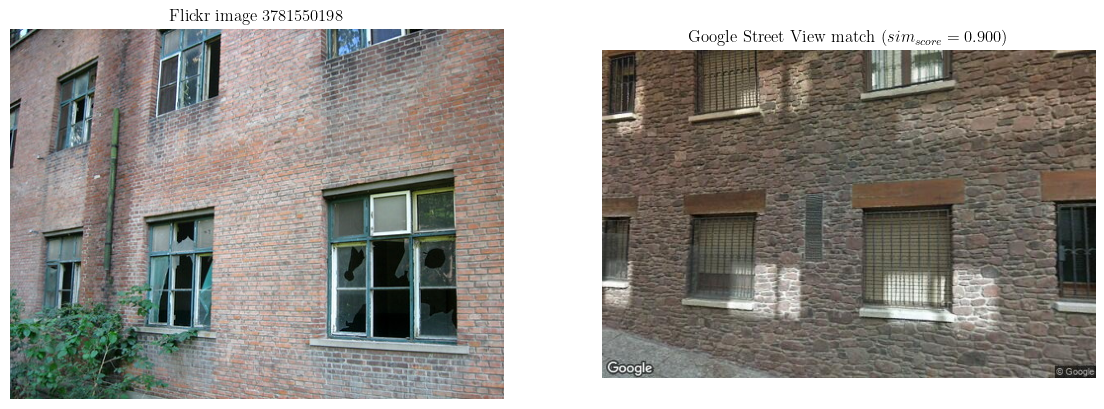}
  \caption{Sample for Flickr image having a $p_{sim}=0.90$}
  \label{fig:sample_flickr_image_0.90_similarity}
\end{figure}

Our analysis is ordered by the appearance of the hyperparameters in the pipeline. First, there is the similarity threshold $t_{sim}$ setting how similar a social media image must be compared to the seed dataset (Figure~\ref{fig:analysis_f1_score_t_sim}). Between 0.70 and 0.80 there is little difference in the resulting F1-score: It is almost constant between 0.50 and 0.52. At the same time the dataset size decreases from 100\% to 2\%. The F1-scores shows a first peak at $t_{sim}=0.83$ with a F1-score of 0.70 and a corresponding dataset size of 0.08\% (23 images in total). For thresholds higher than 0.85 F1-scores become unreliable as the number of images is seven or less. Figures \ref{fig:sample_flickr_image_0.50_similarity}, \ref{fig:sample_flickr_image_0.75_similarity}, and \ref{fig:sample_flickr_image_0.90_similarity} show examples of Flickr images having a $p_{sim}=0.50$, $p_{sim}=0.75$, and $p_{sim}=0.90$.

Figure~\ref{fig:analysis_f1_score_t_score} shows how the F1-score is affected by the object detection score $p_{score}$. The figure starts with $t_{score}=0.30$ because objects with lower scores are not reported by the implementation we used. It increases slightly starting from $t_{score} > 0.30$ with an F1-score of 0.50 up to 0.63 at $t_{score}=0.93$. At the same time, the dataset decreases from 100\% to 3.6\% (this is equal to 930 images). Setting $t_{score} > 0.965$ yields an increase in F1-score to 0.70, but with only 0.2\% or 56 images being considered.

The second parameter from the object detection filtering is $t_{size}$, the minimum size of the \textit{building} or \textit{house} to be found in an image (Figure~\ref{fig:analysis_f1_score_t_size}. Using a threshold $t_{size}=0.2$ yields a F1-score of 0.50. Increasing the threshold up to $t_{size}=0.56$ results in a higher F1-score of 0.52, which is the highest possible value. Raising the threshold further on decreases the F1-score down to 0.40 at $t_{size}=1.0$. At the peak of $t_{size}=0.57$ the remaining dataset consists of 11,234 images (44\%).

As a last parameter in the pipeline, there is the distance between a photographer's position and the next building in the compass direction $p_{dist}$. Figure~\ref{fig:analysis_f1_score_t_dist} depicts the F1-score as a function of the distance. Please note that the threshold is an upper limit. Setting $t_{dist}=0.0$ yields a F1-score of 1.0 based on a single image. Increasing to $t_{dist}=2.2$ provides a first realistic value of 0.48 calculated on 4.8\% of the dataset. Raising the threshold further to $t_{dist}=40.32$ results in the highest possible F1-score of 0.52. At this point, 13,999 images, 69\% of the dataset are included. Higher thresholds lead to a slight decrease of the F1-score down to 0.50 at $t_{dist}=222$.

Overall, the hyperparameters do not have too much influence on the prediction quality. We see that the F1-score is mostly stable around 0.5. Just in case of strict thresholds there are some exceptions: e.g. setting $t_{score}=0.965$ yields a F1-score of 0.70. Adjusting thresholds has more effects on the dataset size: On the one hand, fixing too strict thresholds yields a low number of images. On the other hand, this has a significant effect on the runtime of the whole pipeline. The more images a filter step at the very beginning the higher the over all computational time. This tradeoff needs to be taken into account when applying the filtering steps.



\begin{table}[htb]
\centering
\begin{tabularx}{\textwidth}{Xrrr}
\toprule
Filtering Step                          & \#Images   & \% of dataset              & Execution time [s] \\
\midrule
Flickr LCZ42 Dataset                    & 28,818,438 &                   100.00\% &        \\
Similarity filtering                    &  1,635,592 &                     5.68\% & 0.0236 \\
Object detection filtering              &    891,861 &                     3.09\% & 0.6319 \\
Unique location filtering               &    457,670 &                     1.59\% & 0.0002 \\
Image direction filtering               &     88,593 &                     0.31\% & 1.3333 \\
OSM building in line-of-sight           &     73,207 &                     0.25\% & 0.0008 \\
Labeled OSM building in line-of-sight   &     26,381 &                     0.09\% &        \\
\bottomrule
\end{tabularx}
\caption{Number of Images remaining after each Filtering Step when using $t_{sim}=0.70$, $t_{size}=0.2$, $t_{score}=0.3$, and $t_{dist}=250$. Execution time per image sample in seconds}
\label{tab:flickr_filtering_pipeline_results}
\end{table}

Table~\ref{tab:flickr_filtering_pipeline_results} illustrates the number of images remaining after each filtering step when setting $t_{sim}=0.70$, $t_{score}=0.3$, $t_{size}=0.2$ and $t_{dist}=250$. Additionally, it shows how long it takes to process a single image in a filter step in our setup. While the exact times will change with different setups, the relative comparison allows an assessment of the effectiveness.

Similarity filtering reduces the number of remaining images to less than 6\% of the original dataset at high speed. Discarding all images that do not show a house or a building yields 891,861 images (3.09\% of all images in the original dataset). However, this second step on content filtering takes more than 25 times longer than the similarity check.

Making sure that there is not another image from the very same location filters out 743,731 images, which indicates that almost half of all images were manually tagged. Utilizing a spatial index makes this step the fastest of all filtering steps with 0.2 milliseconds; a hundred times faster than the similarity check. Out of the remaining 457,670 images 88,593 come with a compass orientation. This is 0.31\% of the original dataset. As this step requires downloading additional data using the Flickr API, it takes 1.33 seconds. Please note that most of the time is spent on waiting for the next API request to prevent being blocked by the platform (1 second).

Checking if an OSM building footprint is within the line-of-sight kept 73,207 images and limiting this to labeled OSM buildings gave a final number of 26,381 images. This is 0.09\% of the whole dataset. This step makes again use of the spatial index, which results in the second fastest check of all steps.

There can be more than one image per building, especially touristic landmark buildings can be covered by several images. The 6,955 images from our filtering pipeline were mapped to 18,759 buildings. 5,962 of them are \textit{commercial}, 5,138 are \textit{other}, and 7,659 are \textit{residential}.

\subsection{Prediction Results}

\begin{table}[htbp]
\centering
\begin{tabularx}{\textwidth}{XX|rrr|rrr}
\toprule
                &           &  \multicolumn{3}{c}{Filtered Images} & \multicolumn{3}{c}{Human Validated Images} \\
                & Metric    &  F1 &  Prec   &  Rec   &  F1 &  Prec &  Rec  \\
Architecture    & Class     &           &               &           &           &            &          \\
\midrule
\multirow{3}{*}{densenet121}        & Com   &     0.52 &      0.43 &    0.66    & 0.76  & 0.65  & 0.91  \\
                                    & Oth   &     0.49 &      0.49 &    0.50    & 0.76  & 0.77  & 0.76  \\
                                    & Res   &     0.43 &      0.60 &    0.34    & 0.64  & 0.82  & 0.51  \\
                                    & Avg   &     0.47 &      0.50 &    0.481   & 0.72  & 0.75  & 0.73  \\
\cline{1-8}
\multirow{3}{*}{inceptionv3}        & Com   &     0.51 &      0.44 &    0.61    & 0.76  & 0.68  & 0.91  \\
                                    & Oth   &     0.51 &      0.45 &    0.58    & 0.74  & 0.67  & 0.76  \\
                                    & Res   &     0.39 &      0.63 &    0.28    & 0.55  & 0.87  & 0.51  \\
                                    & Avg   &     0.46 &      0.50 &    0.48    & 0.68  & 0.73  & 0.70  \\
\cline{1-8}
\multirow{3}{*}{mobilenetv2}        & Com   &     0.49 &      0.45 &   0.54  & 0.76  & 0.72  & 0.81  \\
                                    & Oth   &     0.51 &      0.45 &   0.61  & 0.74  & 0.67  & 0.82  \\
                                    & Res   &     0.44 &      0.62 &   0.34  & 0.61  & 0.80  & 0.49  \\
                                    & Avg   &     0.46 &      0.50 &   0.48  & 0.71  & 0.73  & 0.71  \\
\cline{1-8}
\multirow{3}{*}{resnet50v2}         & Com   &     0.48 &      0.43 &   0.55  & 0.73  & 0.67  & 0.81  \\
                                    & Oth   &     0.45 &      0.44 &   0.46  & 0.69  & 0.70  & 0.68  \\
                                    & Res   &     0.45 &      0.53 &   0.40  & 0.58  & 0.65  & 0.53  \\
                                    & Avg   &     0.45 &      0.45 &   0.45  & 0.67  & 0.67  & 0.67  \\
\cline{1-8}
\multirow{3}{*}{vgg16}              & Com   &     0.53 &      0.45 &   0.64  & 0.74  & 0.64  & 0.86  \\
                                    & Oth   &     0.35 &      0.58 &   0.25  & 0.61  & 0.87  & 0.47  \\
                                    & Res   &     0.55 &      0.52 &   0.58  & 0.68  & 0.62  & 0.74  \\
                                    & Avg   &     0.47 &      0.49 &   0.47  & 0.67  & 0.72  & 0.68  \\
\cline{1-8}
\multirow{3}{*}{xception}           & Com   &     0.53 &      0.42 &   0.69  & 0.75  & 0.64  & 0.91  \\
                                    & Oth   &     0.48 &      0.48 &   0.49  & 0.70  & 0.69  & 0.71  \\
                                    & Res   &     0.40 &      0.63 &   0.29  & 0.62  & 0.85  & 0.48  \\
                                    & Avg   &     0.46 &      0.50 &   0.47  & 0.69  & 0.73  & 0.70  \\
\bottomrule
\end{tabularx}
\caption{Prediction results of fine-tuned Google Street View models on filtered Flickr images and on human validated subset. Class labels are abbreviated as highlighted in bold: \textit{\textbf{Com}mercial}, \textit{\textbf{Oth}er}, \textit{\textbf{Res}idential}, and \textit{Avg} stands for the weighted average based on the number of samples}.
\label{tab:flickr_lcz42_new-global_gsv}
\end{table}

Table~\ref{tab:flickr_lcz42_new-global_gsv} summarizes the performance of all fine-tuned models on an image level. Class-wise they behave similar with higher recall values on \textit{commercial} and \textit{other} labeled images and a higher precision value for the \textit{residential} class. One exception in this pattern is VGG16, which has the highest precision score for \textit{other}. The mean F1-score for \textit{commercial} is 0.51, which is slightly higher than the F1-score for \textit{other}, 0.47, and \textit{residential}, 0.37.

\textit{Residential} buildings can appear as single detached houses, townhouses, apartment blocks, or skyscrapers. While the first two forms of residential buildings are easy to predict, the latter ones can be easily confused with the other two classes. This is one possible explanation why we see a high precision for \textit{residential} buildings, but a lower recall.

All architectures show a similar performance on the social media dataset. With respect to the weighted average, the Densenet121 and VGG model show the best F1-score of 0.47, but worst model has a F1-score of 0.45 (Resnet50). Hence, the prediction errors are not model specific, but rather a data issue. Two possible explanations for this behaviour can be either a domain shift when moving from Google Street View images to social media images or a data quality issues in the labels. To investigate the effect of OSM labels on the classification performance, we asked humans to confirm or disprove these labels.

\subsection{Label Verification Results}

For this experiment we selected a random subset of 1,500 social media images with OSM labels, 500 from each class, to be validated by humans. As we required three votes for each image, we got a sense of how difficult the task is for a person seeing only the image and the label. Out of 1,500 images 756 images have full agreement on their label, 744 received inconsistent votes.

\begin{figure}[htb]
  \centering
  \includegraphics[width=0.5\textwidth]{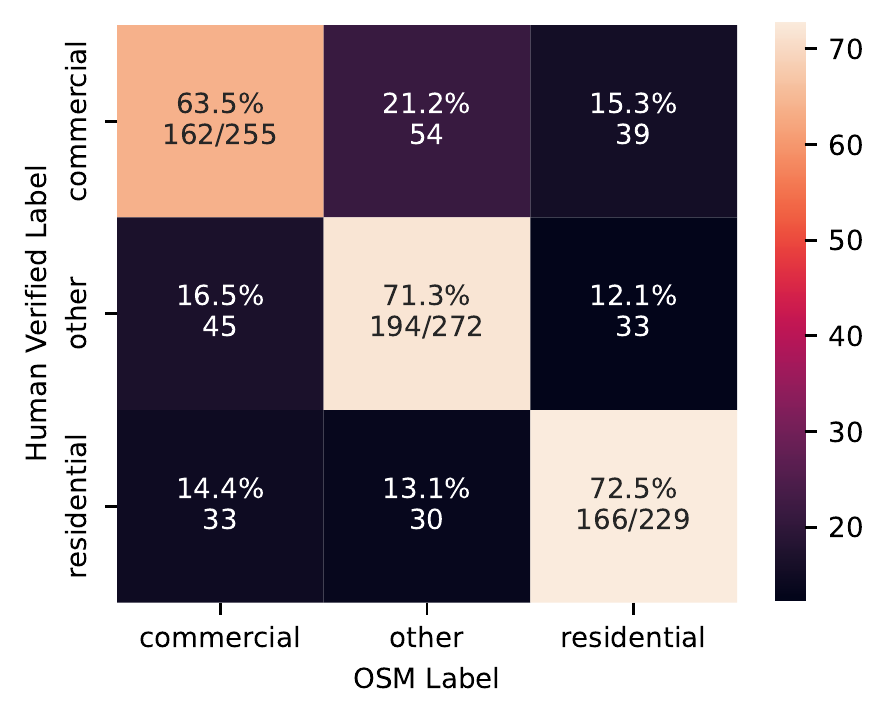}
  \caption{Results from human validation of OSM labels as confusion matrix with only full agreement of human voters}
  \label{fig:confusion_matrix_osm_human_validation}
\end{figure}

Full agreement includes three \textit{unsure} votes as well, so in Figure~\ref{fig:confusion_matrix_osm_human_validation} we focused on the images that received a clear vote, either \textit{correct} or \textit{wrong}. Overall, the accuracy of OSM is 69~\%, but there are subtle differences between the three classes. \textit{Commercial} has 63.5~\% correct labels, which is the lowest value of all classes. On the other hand, \textit{residential} images show 72.5~\% correctness with \textit{other} being similar (71.3~\%).

To assess the true performance of our models we evaluate our models on the subset of images, which received either full agreement on the existing label or a full agreement on a new label. The right part of Figure~\ref{tab:flickr_lcz42_new-global_gsv} shows the F1-score, precision, and recall for this subset. The patterns with respect to precision and recall described above are the same, but all values improved by 0.2. 

Hence, our models yield reasonable results if applied to clear data. In this case, the Densenet121 model yields the best F1-score of 0.72, with MobileNetV2 being second. The VGG16 model is among the worst with a F1-score of 0.67, while it showed up on the first place on the filtered dataset. It seems that the Densenet121 model has best generalized on important features for building function classification.

A big source of error are unclear, mislabeled, or mixed used buildings. Considering that almost have of the images in the human validated subset were did not receive any consistent vote from humans, the performance of these models is sensible.

\section{Discussion}

\subsection{Ambiguity of the Task}
Our simple classification schema of \textit{commercial}, \textit{other}, and \textit{residential} works best if buildings have such a clear function. In historically grown city centers, mixed use building are more common with a retail store on the ground floor and apartments on the upper floors. In these cases our three classes reach their limits and result in an error source. Especially if there is a large sign above the ground floor advertising a retail store on the ground this will likely cause a misclassification.
Additionally, the \textit{other} class is not well defined. Serving as an alternative if none of \textit{commercial} or \textit{residential} truly fit, there are a lot of different patterns pointing to the same final decision. This makes it hard for a CNN to predict this class.

\subsection{Missed Images in Filtering Pipeline}
Although we sampled our Google Street View image dataset on a global scale, there might be still types of buildings that are not covered. This will results in images being discarded although it shows a building and contains valuable geographic information. However, in this case the prediction would likely be wrong in the end as the seed dataset for filtering and the training dataset for fine-tuning are identical. Hence, there would be no benefit in including this image. A possible mitigation could be a more sophisticated sampling algorithm that includes rare building types. 

The same applies to the object detection algorithm. If there are building types that were not in the training dataset of OID, images can be filtered out despite having a building inside, which would be correctly predicted.

\subsection{Correctness of OpenStreetMap Labels}
OSM's primary goal is to provide an open geoinformation service for users to orientate, navigate, and finding places-of-interest (POIs) for their needs. Therefore, \textit{commercial} and \textit{other} buildings, providing any kind of service for a society are more likely mapped than \textit{residential} buildings that have no general purpose. \textit{Residential} buildings are often bulk mapped, so that certain neighborhoods show a high level of completeness, while others do not have any building footprint at all.

Additionally, building functions may change: what used to be a church becomes an apartment building or is abandoned. Validity of labels depends on the activity of OSM's contributors. Hence, in regions with a lot of active contributors, labels will be more up-to-date than in regions with very few contributors.

Last but not least, in areas with few local active contributors like Africa, OSM buildings are mostly mapped by remote users looking at aerial imagery and drawing building polygons accordingly. In such cases, there will most likely be no semantic labels at all.

\subsection{Completeness of OpenStreetMap Building Footprints}
As a VGI platform the completeness of OpenStreetMap buildings polygons varies a lot. If a building footprint in OSM is missing, it can happen that our algorithm assigns an image to a building, which is actually behind the one that it shows.

\subsection{Reference building calculation}
Several images show street view perspectives including more than one building. In such cases, our line-of-sight algorithm will check, which building is the building of the image. Buildings on the left and on the right will be ignored.

\section{Conclusion and Outlook}
\label{sec:conclusion_and_outlook}

In this study we propose a content-first filtering pipeline for social media image datasets to extract geo-spatial information on building functions. By applying five filtering steps, we are able to find relevant images with valid metadata for the given task and relate them to buildings within the line-of-sight. The order of the filter steps ensures scalabilty on large image databases. Moreover, our pipeline has only four hyperparameters for balancing runtime and number of images yielded without strong influence on final prediction results. Based on human validation of our image labels from OSM we show that the limiting performance factor is more the data quality of OSM labels than the models used for predictions. The resulting image dataset with corresponding OSM building IDs and labels are published as a benchmark dataset for urban land use using social media image and weak labels from OSM. Additionally, we provide the human-validated subset with high-quality labels based on three independent votes.

Our pipeline has still many opportunities for refinement. While the cosine similarity measure against a seed dataset ensures fast processing speed, this image retrieval task can be enhanced with more sophisticated algorithms taking different aspects of an image into account \cite{chen2021deep}.
One of the most rigid filtering steps is discarding all images without a compass orientation. Recent approaches that estimate the compass orientation based on aerial imagery could be of help to close this gap \cite{vo2016localizing, regmi2019bridging, shi2020looking}.
As a last step we relate the image to a building using a line-of-sight. Fortunately, the EXIF metadata contains data about the focal length opening a possibility to calculate all buildings that are within the field of view. Based on that an image could be separated into patches with different buildings found during the object detection step. This could yield predictions for many buildings from one image.
Moreover, our classification scheme with \textit{commercial}, \textit{residential}, and \textit{other} focuses on the most crucial classes for population estimation and disaster management. A more fine-grained, multi level scheme could provide more insights into urban development, e.g. education, transportation, health care. Another possible direction could be introducing multi labels to take into account mixed use buildings.

\section*{Acknowledgements}
A big thank you to all anonymous colleagues who helped to verify the image labels from OSM. Without you this study would not be as comprehensive as it is today.

This work is supported by the European Research Council (ERC) under the European Union's Horizon 2020 research and innovation programme (grant agreement number ERC-2016-StG-714087, Acronym: So2Sat, www.so2sat.eu), and Helmholtz Association under the framework of the Young Investigators Group "SiPEO" (VH-NG-1018, https://www.asg.ed.tum.de/sipeo).

\bibliographystyle{elsarticle-harv}
\bibliography{references.bib}

\begin{thebibliography}{44}
\expandafter\ifx\csname natexlab\endcsname\relax\def\natexlab#1{#1}\fi
\expandafter\ifx\csname url\endcsname\relax
  \def\url#1{\texttt{#1}}\fi
\expandafter\ifx\csname urlprefix\endcsname\relax\def\urlprefix{URL }\fi

\bibitem[{Bok{\'a}nyi et~al.(2016)Bok{\'a}nyi, Kondor, Dobos, Seb{\H{o}}k,
  St{\'e}ger, Csabai, and Vattay}]{bokanyi2016race}
Bok{\'a}nyi, E., Kondor, D., Dobos, L., Seb{\H{o}}k, T., St{\'e}ger, J.,
  Csabai, I., Vattay, G., 2016. Race, religion and the city: twitter word
  frequency patterns reveal dominant demographic dimensions in the united
  states. Palgrave Communications 2~(1), 1--9.

\bibitem[{{Camera \& Imaging Products
  Association}(2019)}]{cameraimagingproductsassociationExchangeableImageFile2019}
{Camera \& Imaging Products Association}, May 2019. Exchangeable image file
  formaat for digital still cameras: Exif {{Version}} 2.32.

\bibitem[{Chaudhary et~al.(2019)Chaudhary, D'Aronco, Moy~de Vitry, Leit{\~a}o,
  and Wegner}]{chaudhary2019flood}
Chaudhary, P., D'Aronco, S., Moy~de Vitry, M., Leit{\~a}o, J.~P., Wegner,
  J.~D., 2019. Flood-water level estimation from social media images. ISPRS
  Annals of the Photogrammetry, Remote Sensing and Spatial Information Sciences
  4~(2/W5), 5--12.

\bibitem[{Chen et~al.(2021)Chen, Liu, Wang, Bakker, Georgiou, Fieguth, Liu, and
  Lew}]{chen2021deep}
Chen, W., Liu, Y., Wang, W., Bakker, E., Georgiou, T., Fieguth, P., Liu, L.,
  Lew, M.~S., 2021. Deep image retrieval: A survey. arXiv preprint
  arXiv:2101.11282.

\bibitem[{Chollet(2017)}]{cholletXceptionDeepLearning2017}
Chollet, F., Apr. 2017. Xception: Deep {{Learning}} with {{Depthwise Separable
  Convolutions}}. arXiv:1610.02357 [cs].

\bibitem[{Fang et~al.(2018)Fang, Yuan, Wang, Liu, and Luo}]{fang2018urban}
Fang, F., Yuan, X., Wang, L., Liu, Y., Luo, Z., 2018. Urban land-use
  classification from photographs. IEEE Geoscience and Remote Sensing Letters
  15~(12), 1927--1931.

\bibitem[{Ge et~al.(2018)Ge, Jiang, Xu, Jiang, and
  Ye}]{geExploitingRepresentationsPretrained2018}
Ge, Y., Jiang, S., Xu, Q., Jiang, C., Ye, F., Jul. 2018. Exploiting
  representations from pre-trained convolutional neural networks for
  high-resolution remote sensing image retrieval. Multimedia Tools and
  Applications 77~(13), 17489--17515.

\bibitem[{Gebru et~al.(2017)Gebru, Krause, Wang, Chen, Deng, Aiden, and
  Fei-Fei}]{gebru2017using}
Gebru, T., Krause, J., Wang, Y., Chen, D., Deng, J., Aiden, E.~L., Fei-Fei, L.,
  2017. Using deep learning and google street view to estimate the demographic
  makeup of neighborhoods across the united states. Proceedings of the National
  Academy of Sciences 114~(50), 13108--13113.

\bibitem[{Guttman(1984)}]{guttman1984r}
Guttman, A., 1984. R-trees: A dynamic index structure for spatial searching.
  In: Proceedings of the 1984 ACM SIGMOD international conference on Management
  of data. pp. 47--57.

\bibitem[{Ha et~al.(2018)Ha, Kim, Park, and Kim}]{haImageRetrievalUsing2018}
Ha, I., Kim, H., Park, S., Kim, H., Aug. 2018. Image retrieval using {{BIM}}
  and features from pretrained {{VGG}} network for indoor localization.
  Building and Environment 140, 23--31.

\bibitem[{H{\"a}berle et~al.(2019)H{\"a}berle, Werner, and
  Zhu}]{haberle2019building}
H{\"a}berle, M., Werner, M., Zhu, X.~X., 2019. Building type classification
  from social media texts via geo-spatial textmining. In: IGARSS 2019-2019 IEEE
  International Geoscience and Remote Sensing Symposium. IEEE, pp.
  10047--10050.

\bibitem[{Hamstead et~al.(2018)Hamstead, Fisher, Ilieva, Wood, McPhearson, and
  Kremer}]{hamstead2018geolocated}
Hamstead, Z.~A., Fisher, D., Ilieva, R.~T., Wood, S.~A., McPhearson, T.,
  Kremer, P., 2018. Geolocated social media as a rapid indicator of park
  visitation and equitable park access. Computers, Environment and Urban
  Systems 72, 38--50.

\bibitem[{Havinga et~al.(2021)Havinga, Marcos, Bogaart, Hein, and
  Tuia}]{havinga2021social}
Havinga, I., Marcos, D., Bogaart, P.~W., Hein, L., Tuia, D., 2021. Social media
  and deep learning capture the aesthetic quality of the landscape. Scientific
  reports 11~(1), 1--11.

\bibitem[{He et~al.(2016)He, Zhang, Ren, and Sun}]{heIdentityMappingsDeep2016}
He, K., Zhang, X., Ren, S., Sun, J., Jul. 2016. Identity {{Mappings}} in {{Deep
  Residual Networks}}. arXiv:1603.05027 [cs].

\bibitem[{Hoffmann et~al.(2019)Hoffmann, Wang, Werner, Kang, and
  Zhu}]{hoffmannModelFusionBuilding2019}
Hoffmann, E.~J., Wang, Y., Werner, M., Kang, J., Zhu, X.~X., Jan. 2019. Model
  {{Fusion}} for {{Building Type Classification}} from {{Aerial}} and {{Street
  View Images}}. Remote Sensing 11~(11), 1259.

\bibitem[{Huang et~al.(2018{\natexlab{a}})Huang, Zhao, and
  Song}]{huang2018urban}
Huang, B., Zhao, B., Song, Y., 2018{\natexlab{a}}. Urban land-use mapping using
  a deep convolutional neural network with high spatial resolution
  multispectral remote sensing imagery. Remote Sensing of Environment 214,
  73--86.

\bibitem[{Huang et~al.(2018{\natexlab{b}})Huang, Liu, {van der Maaten}, and
  Weinberger}]{huangDenselyConnectedConvolutional2018}
Huang, G., Liu, Z., {van der Maaten}, L., Weinberger, K.~Q., Jan.
  2018{\natexlab{b}}. Densely {{Connected Convolutional Networks}}.
  arXiv:1608.06993 [cs].

\bibitem[{Huang et~al.(2017)Huang, Rathod, Sun, Zhu, Korattikara, Fathi,
  Fischer, Wojna, Song, Guadarrama, and
  Murphy}]{huangSpeedAccuracyTradeoffs2017}
Huang, J., Rathod, V., Sun, C., Zhu, M., Korattikara, A., Fathi, A., Fischer,
  I., Wojna, Z., Song, Y., Guadarrama, S., Murphy, K., Apr. 2017.
  Speed/accuracy trade-offs for modern convolutional object detectors.
  arXiv:1611.10012 [cs].

\bibitem[{Huang et~al.(2018{\natexlab{c}})Huang, Taubenb{\"o}ck, Mou, and
  Zhu}]{huang2018classification}
Huang, R., Taubenb{\"o}ck, H., Mou, L., Zhu, X.~X., 2018{\natexlab{c}}.
  Classification of settlement types from tweets using lda and lstm. In: IGARSS
  2018-2018 IEEE International Geoscience and Remote Sensing Symposium. IEEE,
  pp. 6408--6411.

\bibitem[{Kang et~al.(2018)Kang, K{\"o}rner, Wang, Taubenb{\"o}ck, and
  Zhu}]{kangBuildingInstanceClassification2018}
Kang, J., K{\"o}rner, M., Wang, Y., Taubenb{\"o}ck, H., Zhu, X.~X., Nov. 2018.
  Building instance classification using street view images. ISPRS Journal of
  Photogrammetry and Remote Sensing 145, 44--59.

\bibitem[{Kruspe et~al.(2021)Kruspe, H{\"a}berle, Hoffmann, Rode-Hasinger,
  Abdulahhad, and Zhu}]{kruspe2021changes}
Kruspe, A., H{\"a}berle, M., Hoffmann, E.~J., Rode-Hasinger, S., Abdulahhad,
  K., Zhu, X.~X., 2021. Changes in twitter geolocations: Insights and
  suggestions for future usage. arXiv preprint arXiv:2108.12251.

\bibitem[{Kuznetsova et~al.(2020)Kuznetsova, Rom, Alldrin, Uijlings, Krasin,
  {Pont-Tuset}, Kamali, Popov, Malloci, Kolesnikov, Duerig, and
  Ferrari}]{kuznetsovaOpenImagesDataset2020}
Kuznetsova, A., Rom, H., Alldrin, N., Uijlings, J., Krasin, I., {Pont-Tuset},
  J., Kamali, S., Popov, S., Malloci, M., Kolesnikov, A., Duerig, T., Ferrari,
  V., Jul. 2020. The {{Open Images Dataset V4}}: Unified image classification,
  object detection, and visual relationship detection at scale. International
  Journal of Computer Vision 128~(7), 1956--1981.

\bibitem[{Langemeyer et~al.(2018)Langemeyer, Calcagni, and
  Baro}]{langemeyer2018mapping}
Langemeyer, J., Calcagni, F., Baro, F., 2018. Mapping the intangible: Using
  geolocated social media data to examine landscape aesthetics. Land use policy
  77, 542--552.

\bibitem[{Laumer et~al.(2020)Laumer, Lang, van Doorn, Mac~Aodha, Perona, and
  Wegner}]{laumer2020geocoding}
Laumer, D., Lang, N., van Doorn, N., Mac~Aodha, O., Perona, P., Wegner, J.~D.,
  2020. Geocoding of trees from street addresses and street-level images. ISPRS
  Journal of Photogrammetry and Remote Sensing 162, 125--136.

\bibitem[{Leichtle et~al.(2019)Leichtle, Lakes, Zhu, and
  Taubenb{\"o}ck}]{leichtleHasDongyingDeveloped2019}
Leichtle, T., Lakes, T., Zhu, X.~X., Taubenb{\"o}ck, H., Nov. 2019. Has
  {{Dongying}} developed to a ghost city? - {{Evidence}} from multi-temporal
  population estimation based on {{VHR}} remote sensing and census counts.
  Computers, Environment and Urban Systems 78, 101372.

\bibitem[{Leung and Newsam(2012)}]{leung2012exploring}
Leung, D., Newsam, S., 2012. Exploring geotagged images for land-use
  classification. In: Proceedings of the ACM multimedia 2012 workshop on
  Geotagging and its applications in multimedia. pp. 3--8.

\bibitem[{Li et~al.(2010)Li, Su, Li, and P~Xing}]{li2010object}
Li, L.-J., Su, H., Li, F.-F., P~Xing, E., 2010. Object bank: A high-level image
  representation for scene classification \& semantic feature sparsification.

\bibitem[{Liu et~al.(2019)Liu, Wang, Wang, Zhang, and
  Lin}]{liuIntelligentSecureContentBased2019}
Liu, F., Wang, Y., Wang, F.-C., Zhang, Y.-Z., Lin, J., 2019. Intelligent and
  {{Secure Content}}-{{Based Image Retrieval}} for {{Mobile Users}}. IEEE
  Access 7, 119209--119222.

\bibitem[{Movshovitz-Attias et~al.(2015)Movshovitz-Attias, Yu, Stumpe, Shet,
  Arnoud, and Yatziv}]{movshovitz2015ontological}
Movshovitz-Attias, Y., Yu, Q., Stumpe, M.~C., Shet, V., Arnoud, S., Yatziv, L.,
  2015. Ontological supervision for fine grained classification of street view
  storefronts. In: Proceedings of the IEEE Conference on Computer Vision and
  Pattern Recognition. pp. 1693--1702.

\bibitem[{Regmi and Shah(2019)}]{regmi2019bridging}
Regmi, K., Shah, M., 2019. Bridging the domain gap for ground-to-aerial image
  matching. In: Proceedings of the IEEE/CVF International Conference on
  Computer Vision. pp. 470--479.

\bibitem[{Ren et~al.(2016)Ren, He, Girshick, and
  Sun}]{renFasterRCNNRealTime2016}
Ren, S., He, K., Girshick, R., Sun, J., Jan. 2016. Faster {{R}}-{{CNN}}:
  Towards {{Real}}-{{Time Object Detection}} with {{Region Proposal Networks}}.
  arXiv:1506.01497 [cs].

\bibitem[{Russakovsky et~al.(2015)Russakovsky, Deng, Su, Krause, Satheesh, Ma,
  Huang, Karpathy, Khosla, Bernstein, Berg, and
  {Fei-Fei}}]{russakovskyImageNetLargeScale2015}
Russakovsky, O., Deng, J., Su, H., Krause, J., Satheesh, S., Ma, S., Huang, Z.,
  Karpathy, A., Khosla, A., Bernstein, M., Berg, A.~C., {Fei-Fei}, L., Dec.
  2015. {{ImageNet Large Scale Visual Recognition Challenge}}. International
  Journal of Computer Vision 115~(3), 211--252.

\bibitem[{Salem et~al.(2020)Salem, Workman, and Jacobs}]{salem2020learning}
Salem, T., Workman, S., Jacobs, N., 2020. Learning a dynamic map of visual
  appearance. In: Proceedings of the IEEE/CVF Conference on Computer Vision and
  Pattern Recognition. pp. 12435--12444.

\bibitem[{Sandler et~al.(2019)Sandler, Howard, Zhu, Zhmoginov, and
  Chen}]{sandlerMobileNetV2InvertedResiduals2019}
Sandler, M., Howard, A., Zhu, M., Zhmoginov, A., Chen, L.-C., Mar. 2019.
  {{MobileNetV2}}: Inverted {{Residuals}} and {{Linear Bottlenecks}}.
  arXiv:1801.04381 [cs].

\bibitem[{Shi et~al.(2020)Shi, Yu, Campbell, and Li}]{shi2020looking}
Shi, Y., Yu, X., Campbell, D., Li, H., 2020. Where am i looking at? joint
  location and orientation estimation by cross-view matching. In: Proceedings
  of the IEEE/CVF Conference on Computer Vision and Pattern Recognition. pp.
  4064--4072.

\bibitem[{Simonyan and Zisserman(2015)}]{simonyanVeryDeepConvolutional2015}
Simonyan, K., Zisserman, A., Apr. 2015. Very {{Deep Convolutional Networks}}
  for {{Large}}-{{Scale Image Recognition}}. arXiv:1409.1556 [cs].

\bibitem[{Sloan et~al.(2013)Sloan, Morgan, Housley, Williams, Edwards, Burnap,
  and Rana}]{lsx2013}
Sloan, L., Morgan, J., Housley, W., Williams, M., Edwards, A., Burnap, P.,
  Rana, O., 2013. Knowing the tweeters: Deriving sociologically relevant
  demographics from twitter. Sociological Research Online 18~(3), 74--84.
\newline\urlprefix\url{https://doi.org/10.5153/sro.3001}

\bibitem[{Srivastava et~al.(2020)Srivastava, Vargas~Munoz, Lobry, and
  Tuia}]{srivastava2020fine}
Srivastava, S., Vargas~Munoz, J.~E., Lobry, S., Tuia, D., 2020. Fine-grained
  landuse characterization using ground-based pictures: a deep learning
  solution based on globally available data. International Journal of
  Geographical Information Science 34~(6), 1117--1136.

\bibitem[{Szegedy et~al.(2015)Szegedy, Vanhoucke, Ioffe, Shlens, and
  Wojna}]{szegedyRethinkingInceptionArchitecture2015}
Szegedy, C., Vanhoucke, V., Ioffe, S., Shlens, J., Wojna, Z., Dec. 2015.
  Rethinking the {{Inception Architecture}} for {{Computer Vision}}.
  arXiv:1512.00567 [cs].

\bibitem[{Terroso-Saenz and Munoz(2020)}]{terroso2020land}
Terroso-Saenz, F., Munoz, A., 2020. Land use discovery based on volunteer
  geographic information classification. Expert Systems with Applications 140,
  112892.

\bibitem[{Vo and Hays(2016)}]{vo2016localizing}
Vo, N.~N., Hays, J., 2016. Localizing and orienting street views using overhead
  imagery. In: European conference on computer vision. Springer, pp. 494--509.

\bibitem[{Wang et~al.(2018)Wang, Lai, Xu, Liu, and
  Lei}]{wangBeautyProductImage2018}
Wang, Q., Lai, J., Xu, K., Liu, W., Lei, L., Oct. 2018. Beauty {{Product Image
  Retrieval Based}} on {{Multi}}-{{Feature Fusion}} and {{Feature
  Aggregation}}. In: Proceedings of the 26th {{ACM}} International Conference
  on {{Multimedia}}. {{MM}} '18. {Association for Computing Machinery}, {New
  York, NY, USA}, pp. 2063--2067.

\bibitem[{Zhang et~al.(2019)Zhang, Sargent, Pan, Li, Gardiner, Hare, and
  Atkinson}]{zhang2019joint}
Zhang, C., Sargent, I., Pan, X., Li, H., Gardiner, A., Hare, J., Atkinson,
  P.~M., 2019. Joint deep learning for land cover and land use classification.
  Remote sensing of environment 221, 173--187.

\bibitem[{Zhu et~al.(2019)Zhu, Deng, and Newsam}]{zhu2019fine}
Zhu, Y., Deng, X., Newsam, S., 2019. Fine-grained land use classification at
  the city scale using ground-level images. IEEE Transactions on Multimedia
  21~(7), 1825--1838.

\end{thebibliography}




\end{document}